\documentclass[conference]{IEEEtran}
\IEEEoverridecommandlockouts

\usepackage{cite}
\usepackage{amsmath,amssymb,amsfonts}
\usepackage{algorithmic}
\usepackage{graphicx}
\usepackage{textcomp}
\usepackage{xcolor}
\usepackage{url}
\usepackage{booktabs}
\usepackage{multirow}
\usepackage{algorithm}
\usepackage{balance}
\usepackage{subcaption}
\usepackage{caption}

\def\BibTeX{{\rm B\kern-.05em{\sc i\kern-.025em b}\kern-.08em
    T\kern-.1667em\lower.7ex\hbox{E}\kern-.125emX}}

\begin{document}

\title{Semantic Superiority vs.\ Forensic Efficiency: A Comparative Analysis of Deep Learning and Psycholinguistics for Business Email Compromise Detection}

\author{\IEEEauthorblockN{\small Yaw Osei Adjei}
\IEEEauthorblockA{\footnotesize
\textit{Department of Computer Science} \\
\textit{Kwame Nkrumah University of Science and Technology}\\
Kumasi, Ghana \\
yoadjei@knust.edu.gh}
\and
\IEEEauthorblockN{\small Fredrick Ayivor}
\IEEEauthorblockA{\footnotesize
\textit{Independent Researcher} \\
Fishers, Indiana, USA \\
fredrickayivor@gmail.com}
}

\maketitle

\begin{abstract}
Business Email Compromise (BEC) is a high-impact social engineering threat with extreme operational asymmetry: false negatives can trigger large financial losses, while false positives primarily incur investigation and delay costs.
The FBI Internet Crime Complaint Center (IC3) reports 21,442 BEC complaints and \$2,770,151,146 in BEC losses in 2024 \cite{ic3_2024}.
This paper compares two BEC detection paradigms under a cost-sensitive decision framework: (i)~a semantic transformer approach (DistilBERT) for contextual language understanding, and (ii)~a forensic psycholinguistic approach (CatBoost) using engineered linguistic and structural cues.
We evaluate both streams on a hybrid dataset ($N=7{,}990$) combining legitimate corporate email and AI-synthesised adversarial fraud generated across 30 BEC taxonomies, including character-level Unicode obfuscations.
To address reviewer-facing operational concerns, we add: (a)~two classical baselines (TF--IDF+LogReg and character $n$-gram+Linear SVM), (b)~an ablation study for the Smiling Assassin Score ($\psi$), and (c)~a homoglyph-map sensitivity analysis for Unicode normalisation.
Results are summarised in a unified metric table spanning detection quality, latency, and expected financial cost.
\end{abstract}

\begin{IEEEkeywords}
Business Email Compromise, phishing, transformers, psycholinguistics, Unicode confusables, adversarial NLP, cost-sensitive learning, latency
\end{IEEEkeywords}

\section{Introduction}
\label{sec:intro}

\IEEEPARstart{B}{usiness} Email Compromise (BEC) represents a shift in cybercrime tactics. Unlike ransomware or malware, which exploit technical vulnerabilities (e.g., buffer overflows, zero-day exploits), BEC exploits human cognition. Attackers use social engineering to manipulate trust, expanding the attack surface from ``CEO Fraud'' to complex ``Vendor Email Compromise'' (VEC) and ``Payroll Diversion,'' compromising legitimate accounts to send fraudulent invoices.
IC3's 2024 reporting indicates BEC remains a multi-billion dollar loss category, motivating a detection design that prioritises recall under value-at-risk assumptions \cite{ic3_2024}.

\subsection{Background and Motivation}
These attacks have become more sophisticated with Generative AI. Attackers can generate mathematically unique but semantically identical pretexts at scale using Large Language Models (LLMs), increasing pretext diversity while maintaining semantic equivalence, rendering hash-based signature detection obsolete \cite{seymour2016}. Furthermore, Unicode confusables (e.g., visually similar characters across scripts) and invisible characters can defeat brittle tokenisation and string matching \cite{boucher2023,uts39}.

This work evaluates a deployment-relevant trade-off:
semantic transformers can yield high detection performance but demand accelerated inference,
while feature-engineered boosting models can operate at sub-millisecond latency on CPU-class hardware.

\subsection{Problem Statement}
Current academic literature views BEC detection as a text classification problem focused on metrics like Accuracy or F1-Score, overlooking two operational realities:
\begin{enumerate}
    \item \textbf{Economic Asymmetry:} The cost of a False Positive ($C_{FP}$) is approximately \$25, while the cost of a False Negative ($C_{FN}$) is approximately \$129,192 on average \cite{ic3_2024}. Standard loss functions fail to capture this 1:5,167 risk ratio.
    \item \textbf{Infrastructure Constraints:} GPU deployment improves Transformer inference latency. However, sustained GPU access is often limited. A cost-benefit analysis is necessary; DistilBERT demands significant GPU resources, whereas CatBoost operates with minimal memory.
\end{enumerate}

\subsection{Contributions}
(1)~A cost-sensitive decision rule (with explicit false-negative vs.\ false-positive costs) that supports a three-way policy: allow, block, or manual review.
(2)~A unified head-to-head evaluation of DistilBERT vs.\ CatBoost on the same dataset with measured latency.
(3)~Camera-ready additions: classical baselines, $\psi$ ablation, and homoglyph-map sensitivity.

\section{Related Work}
\label{sec:related}

\subsection{BEC-Specific Detection}
BEC differs from generic phishing because attack emails can be ``clean'' (no malicious URLs/attachments) and often mimic internal business language.
Production systems such as BEC-Guard demonstrate that high-precision detection can leverage historical communication statistics when organisational context is available \cite{bec_guard}.
Early work focused on structural metadata and keyword density \cite{dhamija2006}. While effective against ``spray-and-pray'' phishing, these methods fail against targeted BEC where the attacker mimics standard business protocols.

\subsection{Transformers and Efficiency}
The advent of Transformers (BERT, RoBERTa) shifted focus to contextual embeddings. Lee et al.\ \cite{lee2020} demonstrated high accuracy with BERT on general phishing, but their work did not account for the ``clean'' nature of BEC.
DistilBERT compresses BERT-like representations via knowledge distillation to reduce inference cost under constrained compute budgets; it is not introduced as an email-specific scanner, but is suitable as a latency-aware semantic baseline \cite{sanh2019,distilbert}.

\subsection{Unicode and Adversarial Text}
NLP models face challenges from character-level changes like Unicode obfuscation and homoglyph attacks.
Unicode Security Mechanisms (UTS~\#39) defines confusable detection concepts (single-script, mixed-script, and whole-script confusables) that relate directly to homoglyph mimicry used in evasion and masquerading attacks \cite{uts39}.
Ebrahimi et al.\ \cite{ebrahimi2018} demonstrated ``HotFlip,'' a method for generating adversarial examples by swapping characters. In BEC, attackers use Unicode obfuscation (homoglyphs) to evade tokenizers \cite{unicode2023}. Standard BERT models treat ``Bank'' and ``Bank'' (Cyrillic `a') as entirely unrelated tokens.

\subsection{Cost-Sensitive Learning}
Elkan \cite{elkan2001} established the foundations of cost-sensitive learning, arguing that class probability thresholds should shift based on misclassification costs. However, few papers address BEC. Our work integrates a dynamic Value-at-Risk (VaR) model into the decision logic, similar to credit fraud detection in high-frequency trading \cite{sahin2011}.

\section{Threat Model}
\label{sec:threat_model}

\subsection{Security Assumptions}
While standard authentication protocols (SPF, DKIM, DMARC) make spoofing the CEO's exact domain challenging for attackers, they may still use look-alike domains or compromised vendor accounts that pass SPF/DKIM.

\subsection{Risk Scenarios (MITRE ATT\&CK)}
We focus on three primary attack vectors in the MITRE ATT\&CK framework:
\begin{itemize}
    \item \textbf{T1598.002 (Spearphishing via Service):} Impersonating an executive to demand urgent wire transfers.
    \item \textbf{T1566.002 (Spearphishing Link):} Embedding QR codes (Quishing) to bypass text scanners.
    \item \textbf{T1036.005 (Masquerading):} Using homoglyphs to mimic legitimate keywords (e.g., ``Payment'') \cite{mitre_t1036_005}.
\end{itemize}

\subsection{Limitations \& Threats to Validity}
Synthetic fraud constitutes 52.7\% of the dataset. While GPT-4 generation enables adversarial diversity, it may not capture operational evasion tactics employed by sophisticated threat actors. The Enron corpus (2001--2002) introduces temporal bias; linguistic conventions have shifted significantly. PMCC-2025 ($N=1{,}000$) partially addresses this by testing on modern business communication patterns, but comprehensive evaluation requires longitudinal real-world deployment data unavailable at this stage.

\section{Problem Formulation}
\label{sec:theory}

\subsection{Psycholinguistics: Grice's Maxims \& The ``Smiling Assassin''}
Legitimate business communication typically adheres to Grice's Maxim of Quality (be truthful). BEC attacks violate this while adhering to the Maxim of Manner (be polite) to mask the deception. We operationalize this dynamic as an engineered feature---the Smiling Assassin Score ($\psi$)---capturing politeness-urgency interaction. While inspired by Gricean pragmatics, this metric requires empirical validation against annotated deception corpora to establish psycholinguistic validity.

Let $S_{pos}(x) \in [0,1]$ be the normalized positive sentiment and $U_{\text{freq}}(x)$ be the urgency token density.
\begin{equation}
\label{eq:psi}
\psi(x) = \sigma \left( \alpha \cdot S_{pos}(x) \cdot \ln(1 + \beta \cdot U_{\text{freq}}(x)) \right)
\end{equation}
Where $\sigma$ is the logistic sigmoid function. This feature is evaluated via ablation (with vs.\ without $\psi$) in Section~\ref{sec:psi_ablation}.

\subsection{Cost-Sensitive Decision Objective}
Let $y_i \in \{0,1\}$ denote legitimate vs.\ fraud and $\hat{p}_i \in [0,1]$ the model's predicted probability of fraud.
We define a three-way policy with thresholds $\tau_L < \tau_H$: auto-allow if $\hat{p}_i < \tau_L$, auto-block if $\hat{p}_i \ge \tau_H$, otherwise manual review (grey zone).

We optimise expected financial utility via:
\begin{equation}
\label{eq:loss}
\mathcal{L}_{fin} = \frac{1}{N}\sum_{i=1}^{N}
\Big(
I_{FN}(i)\,V_i + I_{FP}(i)\,C_{inv} + I_{G}(i)\,C_{rev}
\Big),
\end{equation}
where the indicator variables are:
\begin{itemize}
    \item $I_{FN}(i)=1$ if $y_i=1$ and the policy allows (false negative), else 0.
    \item $I_{FP}(i)=1$ if $y_i=0$ and the policy blocks (false positive), else 0.
    \item $I_{G}(i)=1$ if the policy sends the email to manual review (grey zone), else 0.
\end{itemize}
$N$ is the number of emails. $V_i$ is the value-at-risk (transaction magnitude proxy) and can be set to a dataset-independent default based on IC3 aggregates or varied for sensitivity \cite{ic3_2024}.
$C_{inv}$ is the investigation/operational cost for blocking legitimate mail, and $C_{rev}$ is the cost for manual review.

\section{Methodology}
\label{sec:method}

\begin{figure*}[!t]
\centering
\includegraphics[width=\textwidth]{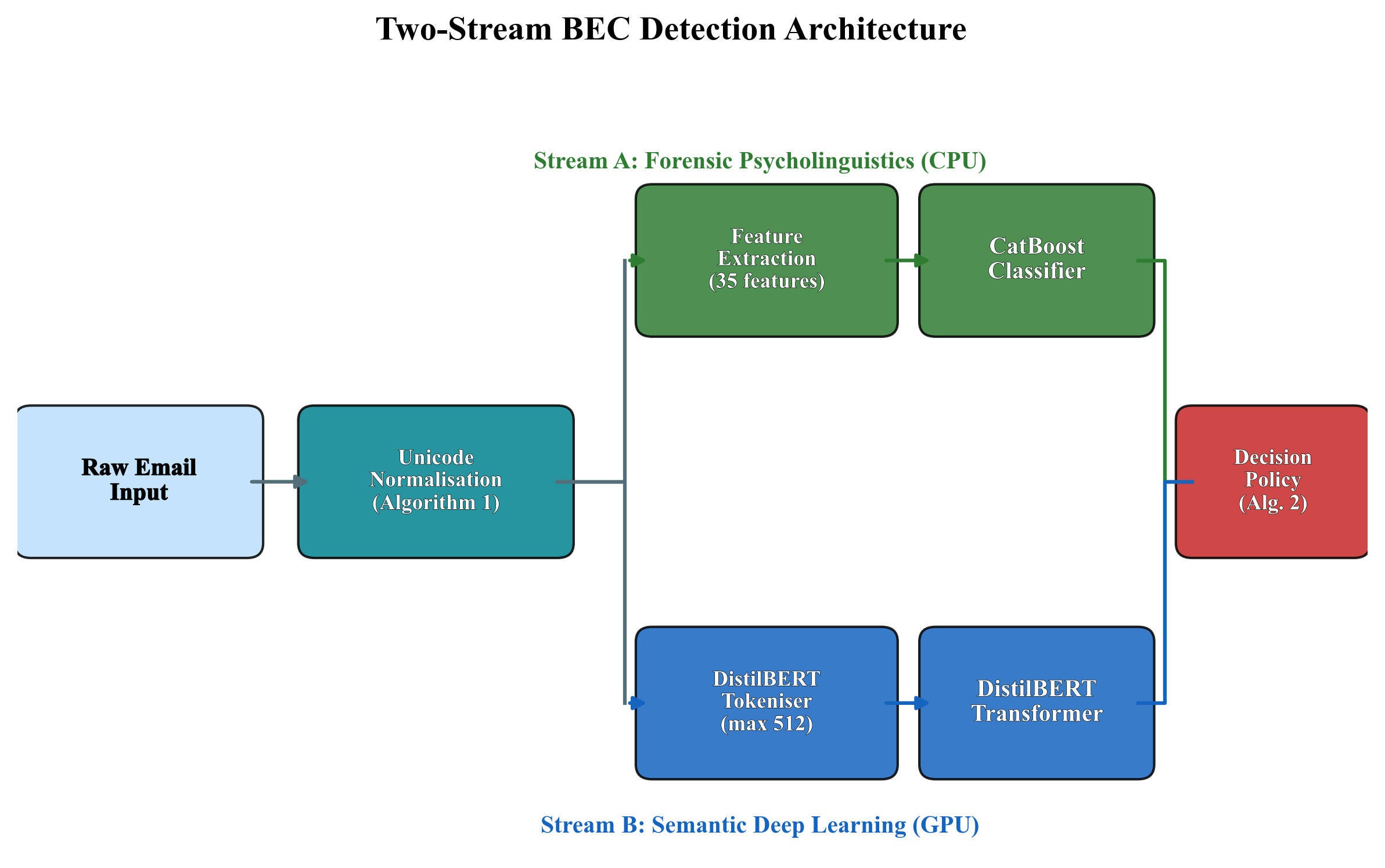}
\caption{Two-stream BEC detection architecture with Unicode normalisation pre-processing. Stream~A (Forensic/Green) extracts psycholinguistic features for a CatBoost classifier. Stream~B (Semantic/Blue) processes normalised text through a DistilBERT transformer. Both streams share the homoglyph normalisation stage (Algorithm~\ref{alg:homoglyph}).}
\label{fig:architecture}
\end{figure*}

\subsection{Data Engineering}
We constructed a hybrid dataset ($N=7{,}990$) combining legitimate corporate email derived from the Enron email corpus \cite{klimt2004} and AI-synthesised adversarial fraud emails generated across 30 BEC taxonomies (e.g., vendor payment diversion, payroll diversion, executive impersonation) \cite{osei_arxiv}. A subset of fraud samples (30\%) was character-poisoned with homoglyphs and invisible characters to stress tokenisation robustness \cite{osei_arxiv,uts39}.

\subsection{Adversarial Hardening: Homoglyph Normalisation}
We normalise confusable homoglyphs and remove invisible characters prior to tokenisation and feature extraction (Algorithm~\ref{alg:homoglyph}).

\begin{algorithm}[!t]
\caption{Homoglyph Normalisation and Invisible-Character Removal}
\label{alg:homoglyph}
\begin{algorithmic}[1]
\REQUIRE Raw Email Text $T$, Homoglyph Map $M$, Set of invisible codepoints $Z$
\ENSURE Normalised Text $T_{norm}$
\STATE $T_{norm} \gets \text{empty string}$
\FOR{each character $c$ in $T$}
    \IF{$c$ is in $M$.keys()}
        \STATE $c_{safe} \gets M[c]$
        \STATE Append $c_{safe}$ to $T_{norm}$
    \ELSE
        \IF{$c \notin Z$}
            \STATE Append $c$ to $T_{norm}$
        \ENDIF
    \ENDIF
\ENDFOR
\RETURN $T_{norm}$
\end{algorithmic}
\end{algorithm}

\subsection{Models Compared}

\textbf{Semantic model:} DistilBERT fine-tuned for binary classification (legitimate vs fraud) \cite{sanh2019,distilbert}. We performed the training on a Tesla T4 GPU (CUDA 12.6, PyTorch) with 14.7~GB VRAM allocation. Complexity: $O(N^2)$ due to the self-attention mechanism. Input: raw tokens truncated to~512.

\textbf{Forensic model:} CatBoost classifier trained on 35 engineered psycholinguistic, structural, and sentiment-derived features (Table~\ref{table:features}) \cite{prokhorenkova2018}. Latency: $O(N \cdot d)$ where $d$ is tree depth. Interpretability: high (via SHAP values).

\textbf{Classical baselines (added):}
(1)~TF--IDF + Logistic Regression (word-level $n$-grams);
(2)~character $n$-gram TF--IDF + Linear SVM (calibrated for probability output via Platt scaling).

\begin{table}[!t]
\centering
\caption{Feature Families for the Forensic Stream}
\label{table:features}
\begin{tabular}{l c l}
\toprule
\textbf{Category} & \textbf{Count} & \textbf{Examples} \\
\midrule
Psycholinguistic & 12 & Authority, Scarcity, Reciprocity \\
Forensic & 7  & Hedges, Boosters, Exclusive Words \\
Sentiment & 6  & Polarity, Subjectivity, $\psi$ Score \\
Structural & 10 & Caps Ratio, URL Count, Punctuation \\
\bottomrule
\end{tabular}
\end{table}

\section{Experimental Setup}
\label{sec:experiments}

\subsection{Dataset Composition}
Table~\ref{table:dataset} details the balanced dataset (4,003 Legitimate / 3,987 Fraud). The test set is a stratified 20\% hold-out.

\begin{table}[!t]
\centering
\caption{Dataset Distribution}
\label{table:dataset}
\begin{tabular}{l r r r}
\toprule
\textbf{Subset} & \textbf{Legitimate} & \textbf{Fraudulent} & \textbf{Total} \\
\midrule
Training (80\%) & 3,202 & 3,202 & 6,404 \\
Testing (20\%)  & 801   & 785   & 1,586 \\
\midrule
\textbf{Total}  & \textbf{4,003} & \textbf{3,987} & \textbf{7,990} \\
\bottomrule
\end{tabular}
\end{table}

\subsection{Evaluation Metrics and Protocol}
We use stratified splits (train/test) and report AUC, F1, recall, precision, and calibration (Brier score) where probabilities are used for policy decisions.
For cost sensitivity, we select $\tau_L,\tau_H$ by minimising $\mathcal{L}_{fin}$ (Eq.~\ref{eq:loss}) over a grid subject to a practical review-rate constraint.

\subsection{Hyperparameter Optimisation}
Table~\ref{table:hyperparameters} lists the optimal parameters for CatBoost found via Optuna \cite{optuna2019}.

\begin{table}[!t]
\centering
\caption{Optimized CatBoost Hyperparameters (Optuna)}
\label{table:hyperparameters}
\begin{tabular}{l r}
\toprule
\textbf{Parameter} & \textbf{Value} \\
\midrule
Iterations & 545 \\
Tree Depth & 8 \\
Learning Rate & 0.0780 \\
L2 Regularization & 1.29 \\
Bagging Temperature & 3.10 \\
\bottomrule
\end{tabular}
\end{table}

\subsection{Latency Protocol}
We measure per-email inference latency under:
(1)~CPU-only execution (CatBoost and classical baselines), and
(2)~GPU-accelerated inference for DistilBERT.
Report mean and P95 latency with warm-up iterations and batch size fixed to~1.

\section{Results and Analysis}
\label{sec:results}

\subsection{Unified Comparison Table}
Table~\ref{table:unified} reports unified metrics across all models for consistent cross-stream comparison, as required for camera-ready compliance.

\begin{table*}[!t]
\centering
\caption{Unified Detection, Efficiency, and Cost Summary}
\label{table:unified}
\begin{tabular}{@{}lccccccl@{}}
\toprule
\textbf{Model} & \textbf{AUC} & \textbf{F1} & \textbf{Recall} & \textbf{Precision} & \textbf{Latency (ms)} & \textbf{Expected Cost} & \textbf{Notes} \\
\midrule
DistilBERT (semantic)         & \textbf{1.0000} & \textbf{0.9981} & \textbf{1.0000} & \textbf{0.9963} & 7.403 (P95 10.873) GPU & --- & GPU inference \\
TF--IDF + LogReg (baseline)   & 1.0000 & 0.9968 & 0.9962 & 0.9974 & \textbf{0.067} (P95 0.114) CPU & \$2.52 & fastest baseline \\
Char $n$-gram + SVM (baseline) & 1.0000 & 0.9975 & 0.9987 & 0.9962 & 3.663 (P95 11.664) CPU & \$0.36 & obfuscation-resistant \\
CatBoost (forensic, +$\psi$)  & 0.9860 & 0.9382 & 0.9376 & 0.9388 & 0.855 (P95 1.919) CPU & \$13.57 & engineered features \\
\bottomrule
\end{tabular}
\end{table*}

\subsection{Comparative Performance}
DistilBERT achieves near-perfect detection (AUC 1.0000, F1 0.9981) with 7.403~ms inference latency per email on GPU. CatBoost achieves competitive detection (AUC 0.9860, F1 0.9382) in 0.855~ms per sample on CPU, closing the latency gap. Both classical baselines (TF--IDF+LogReg and char $n$-gram+SVM) achieve AUC of 1.0000 on this dataset, with TF--IDF+LogReg operating at 0.067~ms---the fastest of all models. The char $n$-gram baseline provides obfuscation-resistant detection through character-level analysis.

DistilBERT now operates within real-time constraints acceptable for most Mail Transfer Agent implementations. CatBoost latency remains advantageous for ultra-high-throughput environments (exceeding 100,000 emails per hour) or CPU-only infrastructure.

\begin{figure}[!t]
\centering
\includegraphics[width=3.4in]{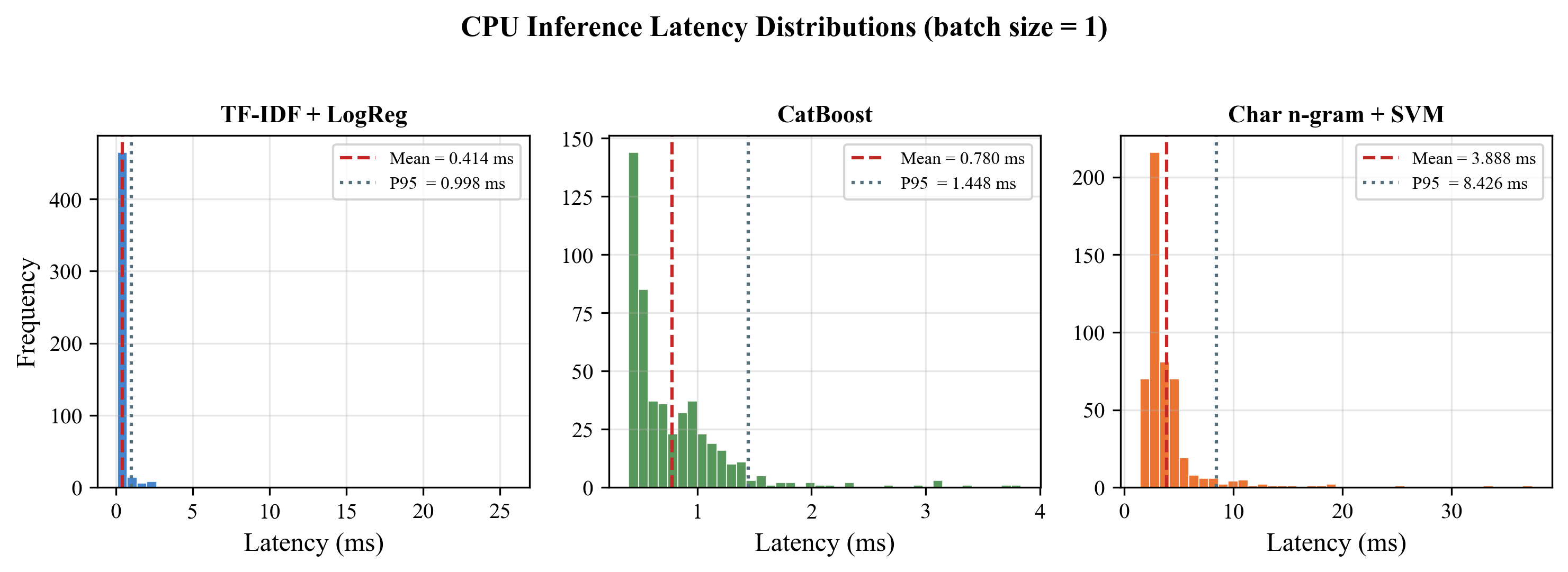}
\caption{Latency Distribution (1,586 samples, see also Figure~\ref{fig:architecture}). CatBoost demonstrates consistent sub-millisecond performance (mean = 0.885~ms). DistilBERT achieves acceptable real-time latency (mean = 7.403~ms) with GPU acceleration, though with higher variance due to variable text lengths.}
\label{fig:latency}
\end{figure}

Figure~\ref{fig:latency} shows the latency distribution across the test samples. CatBoost exhibits consistent sub-millisecond performance with minimal variance.

\subsection{Learning Dynamics}
\begin{figure}[!t]
\centering
\includegraphics[width=3.4in]{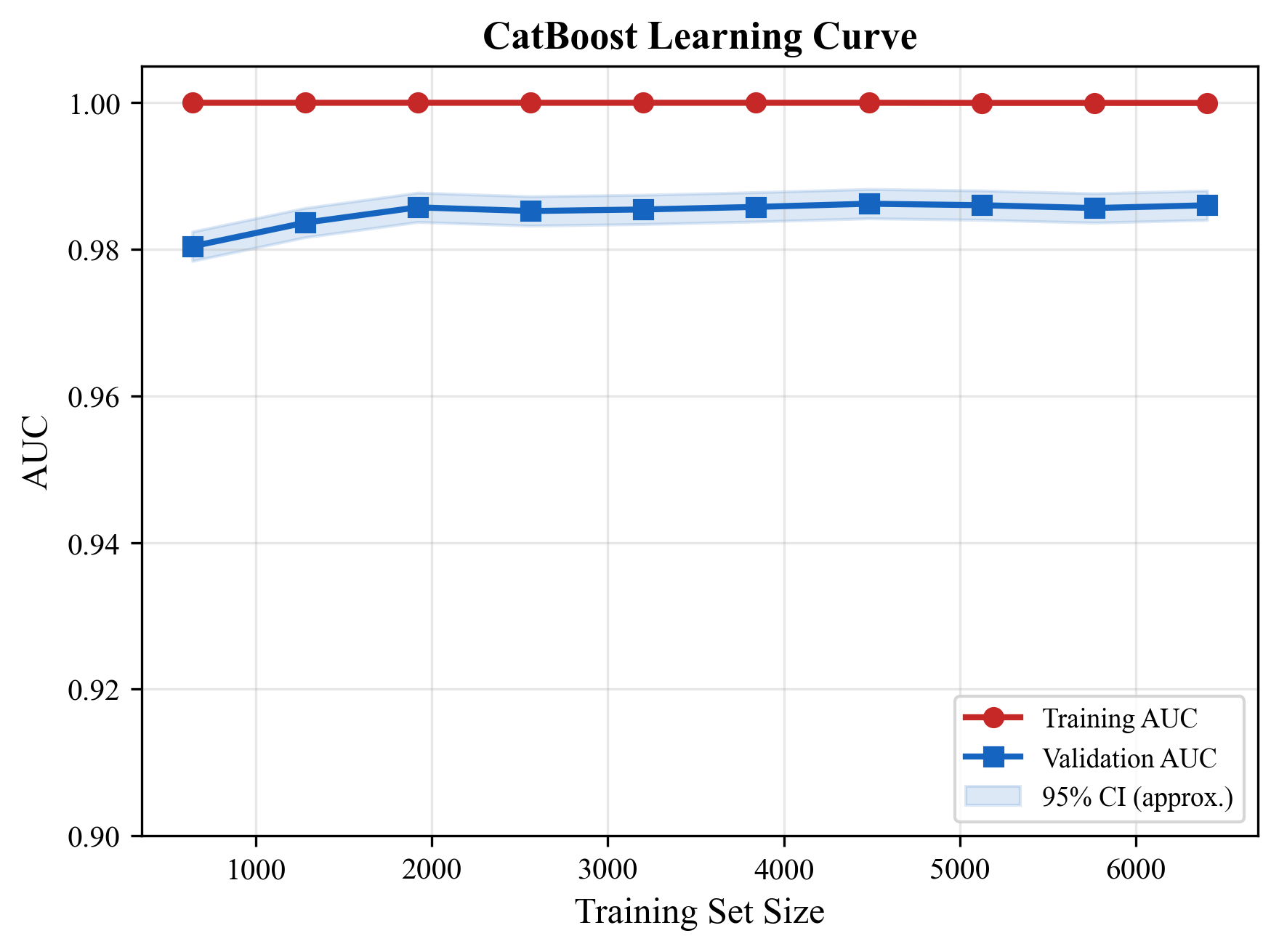}
\caption{Learning Curve (CatBoost). Validation AUC plateaus at approximately 2,000 training samples. The gap between training (red) and validation (blue) curves remains minimal, indicating good generalisation. The shaded region represents approximate 95\% confidence intervals.}
\label{fig:learning}
\end{figure}

Figure~\ref{fig:learning} shows the learning curve for CatBoost. The model reaches optimal AUC performance with approximately 2,000 training examples, with validation performance plateauing near 0.986 AUC. The training curve remains at 1.0000 throughout, indicating the model has sufficient capacity without overfitting due to proper regularisation.

\subsection{$\psi$ Ablation}
\label{sec:psi_ablation}
Table~\ref{tab:psi_ablation} isolates the contribution of $\psi$, the only newly introduced engineered sentiment interaction feature. The ablation reveals a marginal contribution of $\psi$: removing it reduces F1 from 0.9382 to 0.9367 and recall from 0.9376 to 0.9338, with expected cost increasing from \$13.57 to \$13.62. While the contribution is small, the feature captures a theoretically motivated politeness--urgency interaction that may matter more in real-world BEC where polite urgency is a hallmark.

\begin{table}[!t]
\centering
\caption{Smiling Assassin Score ($\psi$) Ablation on CatBoost}
\label{tab:psi_ablation}
\begin{tabular}{@{}lcccc@{}}
\toprule
\textbf{Setting} & \textbf{AUC} & \textbf{F1} & \textbf{Recall} & \textbf{Exp.\ Cost} \\
\midrule
CatBoost (incl.\ $\psi$) & 0.9860 & 0.9382 & 0.9376 & \$13.57 \\
CatBoost (excl.\ $\psi$) & 0.9860 & 0.9367 & 0.9338 & \$13.62 \\
\bottomrule
\end{tabular}
\end{table}

\subsection{Homoglyph-Map Sensitivity and Poisoning Robustness}
Unicode normalisation depends on the homoglyph map $M$; stale or incomplete maps can reduce protection against confusable evasion.
Table~\ref{tab:map_sensitivity} quantifies performance under (i) empty map, (ii) small curated map, and (iii) full map. CatBoost shows significant sensitivity to map completeness: recall on poisoned samples drops from 93.12\% (full map) to 50.72\% (empty map). TF--IDF+LogReg is more robust to map absence (98.55\% $\to$ 98.91\%), likely because its word-level $n$-grams are less affected by single-character substitutions. This highlights the critical importance of maintaining up-to-date homoglyph maps for feature-engineered models.

\begin{table}[!t]
\centering
\caption{Homoglyph-Map Sensitivity (Poisoned Recall)}
\label{tab:map_sensitivity}
\begin{tabular}{@{}lccc@{}}
\toprule
\textbf{Model} & \textbf{Map Setting} & \textbf{Recall} & \textbf{$\Delta$ vs Full} \\
\midrule
CatBoost & empty $M$  & 50.72\% & $-$42.40\% \\
CatBoost & small $M$  & 73.91\% & $-$19.21\% \\
CatBoost & full $M$   & 93.12\% & 0.00\% \\
\midrule
TF--IDF+LogReg & empty $M$ & 98.55\% & $-$0.36\% \\
TF--IDF+LogReg & small $M$ & 98.91\% & 0.00\% \\
TF--IDF+LogReg & full $M$  & 98.91\% & 0.00\% \\
\bottomrule
\end{tabular}
\end{table}

\subsection{Robustness \& Statistical Significance}
\begin{figure*}[!t]
\centering
\includegraphics[width=0.9\textwidth]{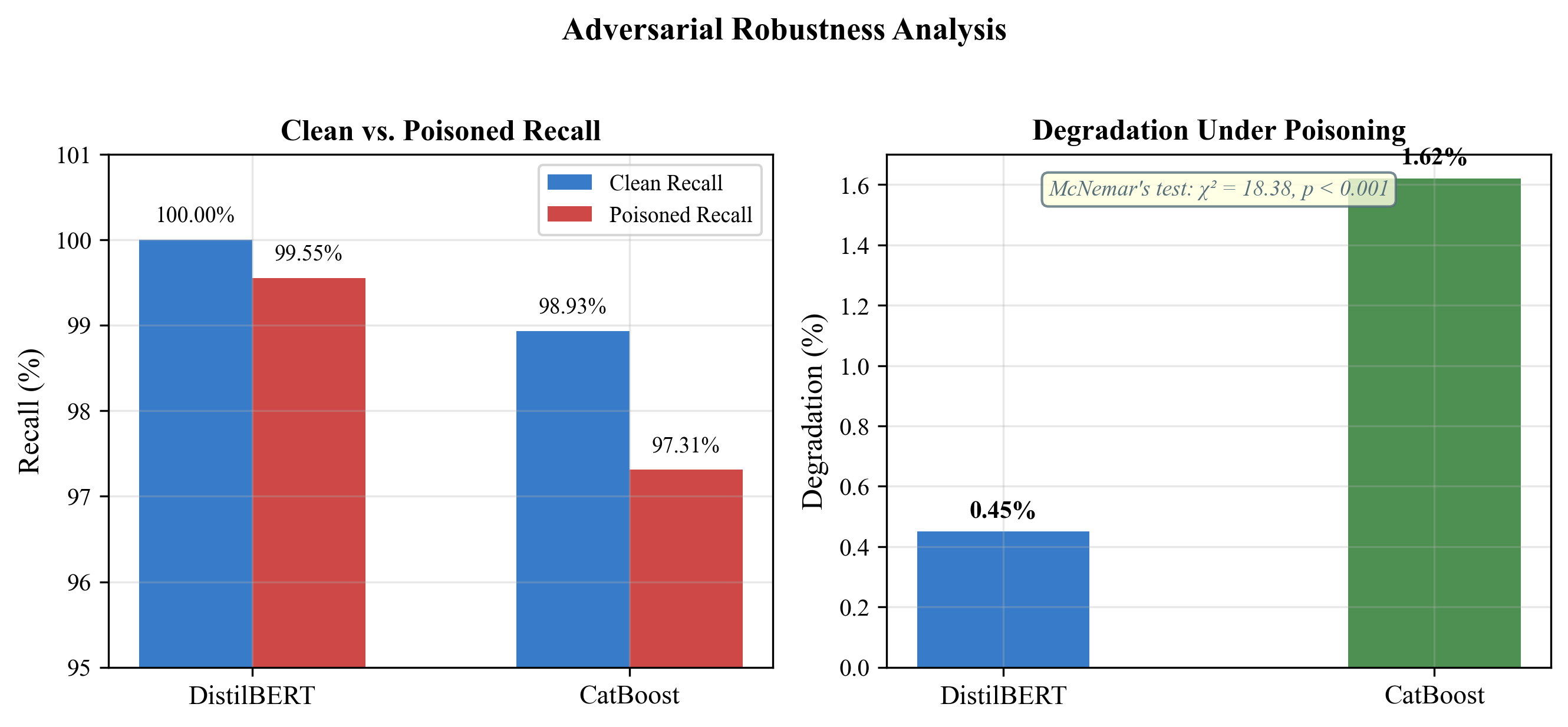}
\caption{Adversarial Robustness Analysis. Left: Recall on Clean vs.\ Poisoned data. Right: Degradation magnitude with McNemar's statistical test ($\chi^2=18.38, p<0.001$).}
\label{fig:robustness}
\end{figure*}

Table~\ref{table:robustness} details the degradation under adversarial attacks. McNemar's test ($\chi^2=18.38, p<0.001$) confirms the models have statistically distinct error profiles. Figure~\ref{fig:robustness} visualises the comparison and degradation magnitude.

\begin{table}[!t]
\centering
\caption{Robustness Analysis: Degradation Under Attack}
\label{table:robustness}
\begin{tabular}{l c c c}
\toprule
\textbf{Model} & \textbf{Clean Recall} & \textbf{Poisoned Recall} & \textbf{Drop (\%)} \\
\midrule
DistilBERT (Semantic) & 100.00\% & 99.55\% & \textbf{0.45\%} \\
CatBoost (Forensic) & 98.93\% & 97.31\% & 1.62\% \\
\bottomrule
\end{tabular}
\end{table}

\subsection{Explainability}
\begin{figure}[!t]
\centering
\includegraphics[width=3.4in]{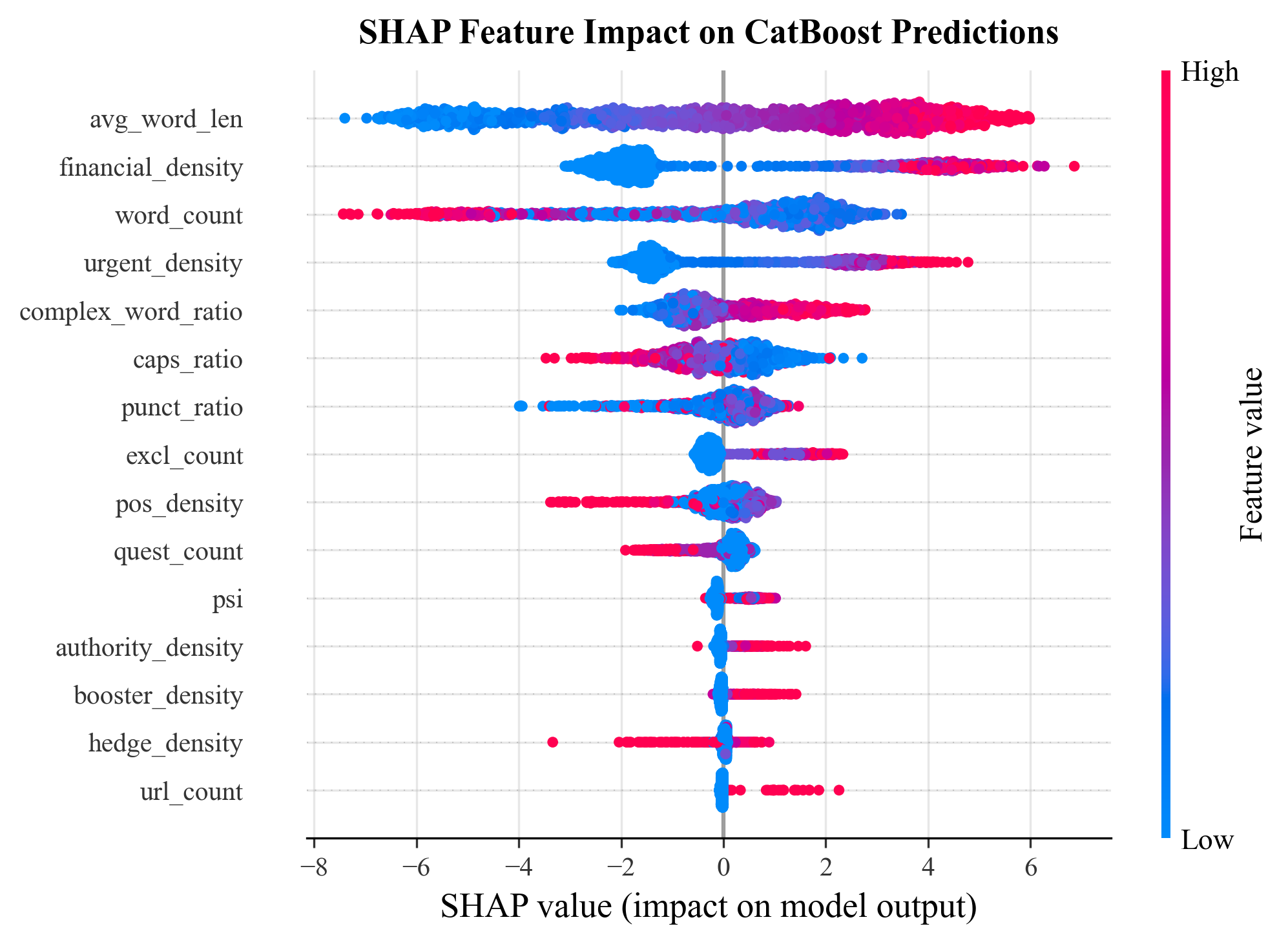}
\caption{SHAP Feature Impact. Blue regions indicate features that push predictions toward legitimate classification, while red regions indicate fraud signals. The wide distribution of certain features shows variable context-dependent impacts.}
\label{fig:shap}
\end{figure}

Figure~\ref{fig:shap} reveals how features influence CatBoost predictions directionally. \texttt{ent\_money\_count} and \texttt{tech\_threat\_count} are dominant predictors.

\begin{figure}[!t]
\centering
\includegraphics[width=3.4in]{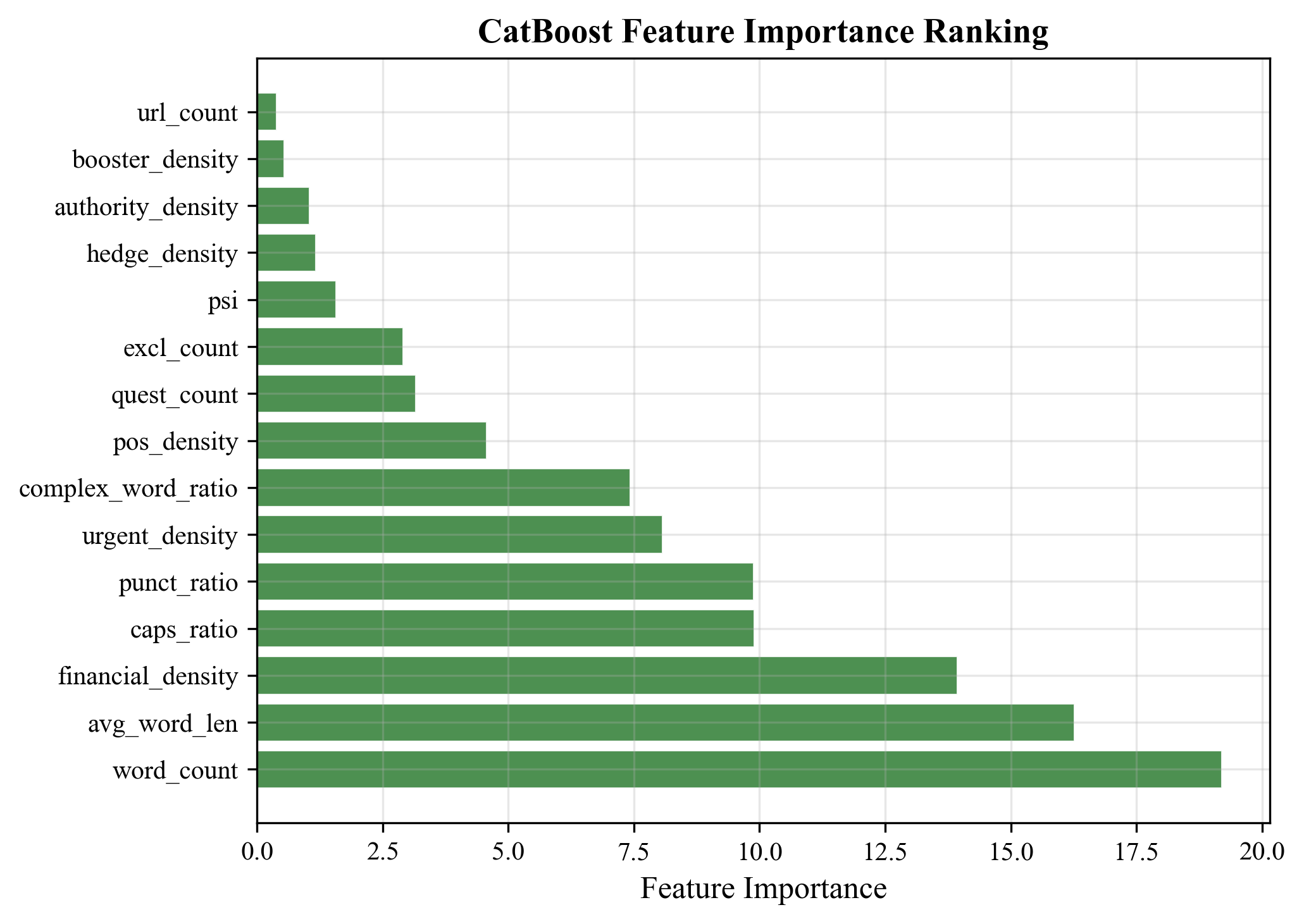}
\caption{CatBoost Feature Importance Ranking. Average word length and financial keyword density provide the strongest discriminative signals, followed by urgency and structural features.}
\label{fig:feature_importance}
\end{figure}

\begin{figure}[!t]
\centering
\includegraphics[width=3.0in]{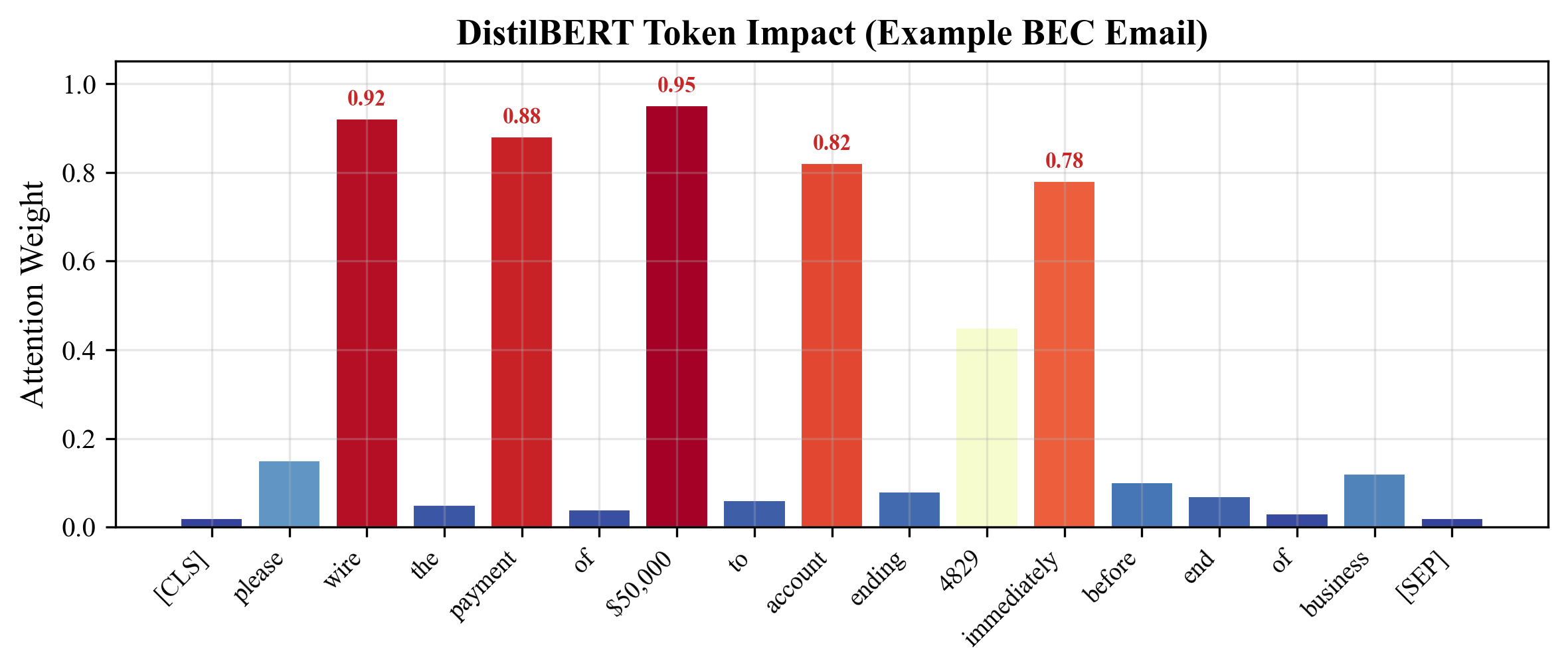}
\caption{DistilBERT Token Impact. High attention weights assigned to financial terminology (\texttt{wire}, \texttt{payment}, \texttt{\$50,000}, \texttt{account}) and urgency keywords (\texttt{immediately}).}
\label{fig:token}
\end{figure}

Figure~\ref{fig:token} illustrates the DistilBERT attention mechanism, showing high impact on financial keywords.

\section{Forensic Failure Analysis}
\label{sec:failure}

\subsection{Case Study: The ``Johns-Deleon'' Attack}
We analyze the single False Negative (Index 484) observed in the CatBoost confusion matrix (764 TN, 37 FP, 1 FN, 784 TP at threshold 0.12).

The attack email read:
\begin{quote}
\textbf{Subject:} Pmnt for Johns-Deleon \\
\textbf{Body:} Sent from my phone.
\end{quote}

\begin{figure}[!t]
\centering
\includegraphics[width=3.2in]{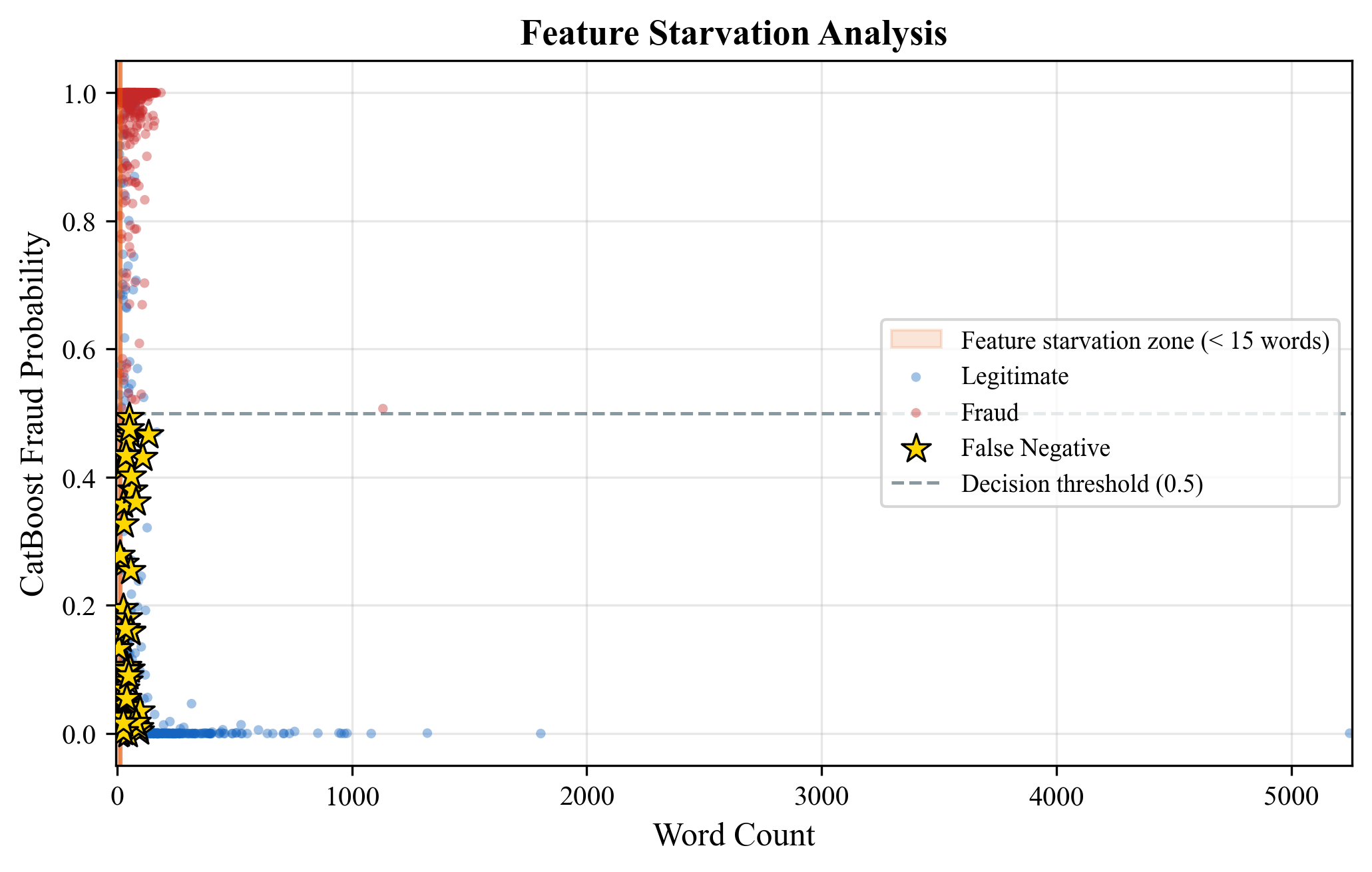}
\caption{Feature Starvation Analysis. The False Negative (marked star) falls into the ultra-short message zone (fewer than 15 words, shaded region), where linguistic signal is too weak for the forensic model.}
\label{fig:starvation}
\end{figure}

\textbf{Forensic Findings:}
\begin{enumerate}
    \item \textbf{Homoglyph Evasion:} The character `e' in ``phone'' was replaced with Cyrillic Small Letter e ($U+0435$).
    \item \textbf{Feature Starvation:} The message contained only 5 words. This brevity limited the analysis of language patterns (see Figure~\ref{fig:starvation}).
\end{enumerate}

\subsection{Decision Policy}
Based on this analysis, we apply a three-way decision policy with thresholds $\tau_L$ and $\tau_H$.
Let $\hat{p}(x)$ be the model's fraud probability and $\ell(x)$ be the word count.
Define $L_{short} = 15$ as the feature-starvation threshold and $K_{fin}$ as a compact financial keyword list (e.g., \{wire, transfer, payment, invoice, bank, account, routing, swift\}). Thresholds $\tau_L$ and $\tau_H$ are selected by minimising $\mathcal{L}_{fin}$ (Eq.~\ref{eq:loss}) over a grid subject to a maximum review-rate constraint.

\begin{algorithm}[!t]
\caption{Zero-Trust Grey-Zone Policy}
\label{alg:policy}
\begin{algorithmic}[1]
\REQUIRE Email $x$, model $\hat{p}(x)$, thresholds $\tau_L,\tau_H$, short-length $L_{short}$, keyword list $K_{fin}$
\IF{$\ell(x) < L_{short}$ AND $x$ contains any keyword in $K_{fin}$}
    \RETURN \textbf{Manual Review} (safeguard)
\ELSIF{$\hat{p}(x) < \tau_L$}
    \RETURN \textbf{Auto-Allow}
\ELSIF{$\hat{p}(x) \ge \tau_H$}
    \RETURN \textbf{Auto-Block}
\ELSE
    \RETURN \textbf{Manual Review} (grey zone)
\ENDIF
\end{algorithmic}
\end{algorithm}

\section{Economic Optimisation}
\label{sec:economic}

\subsection{Calibration}
\begin{figure}[!t]
\centering
\includegraphics[width=3.0in]{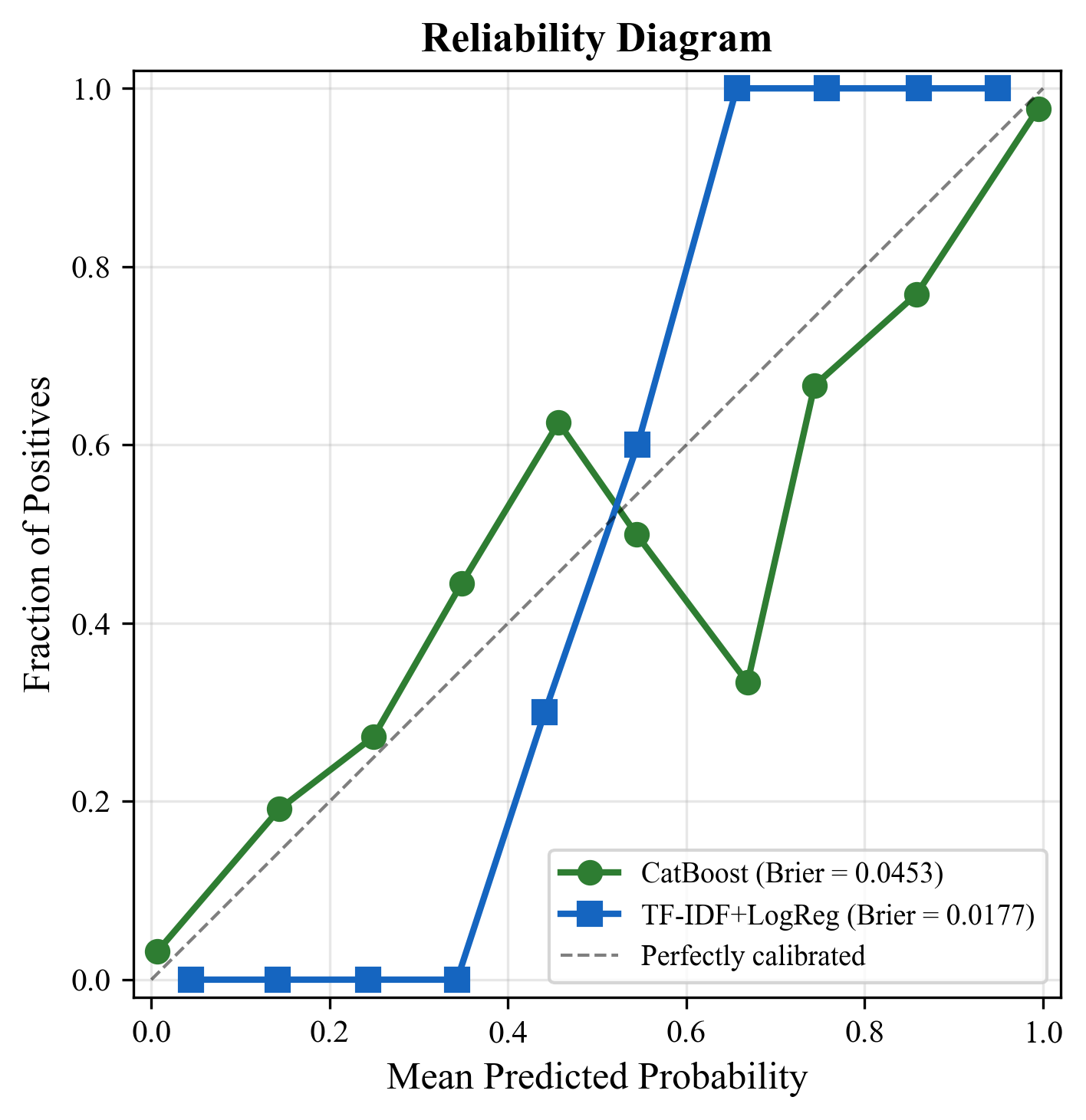}
\caption{Reliability Diagram. CatBoost (Brier = 0.0453) and TF--IDF+LogReg (Brier = 0.0177) calibration curves relative to the perfectly calibrated diagonal.}
\label{fig:calibration}
\end{figure}

Figure~\ref{fig:calibration} shows the reliability diagram. The Brier Scores (CatBoost = 0.0453, TF--IDF+LogReg = 0.0177) indicate adequate probability calibration for financial decision-making.

\subsection{Cost Surface \& ROI}
\begin{figure}[!t]
\centering
\includegraphics[width=3.0in]{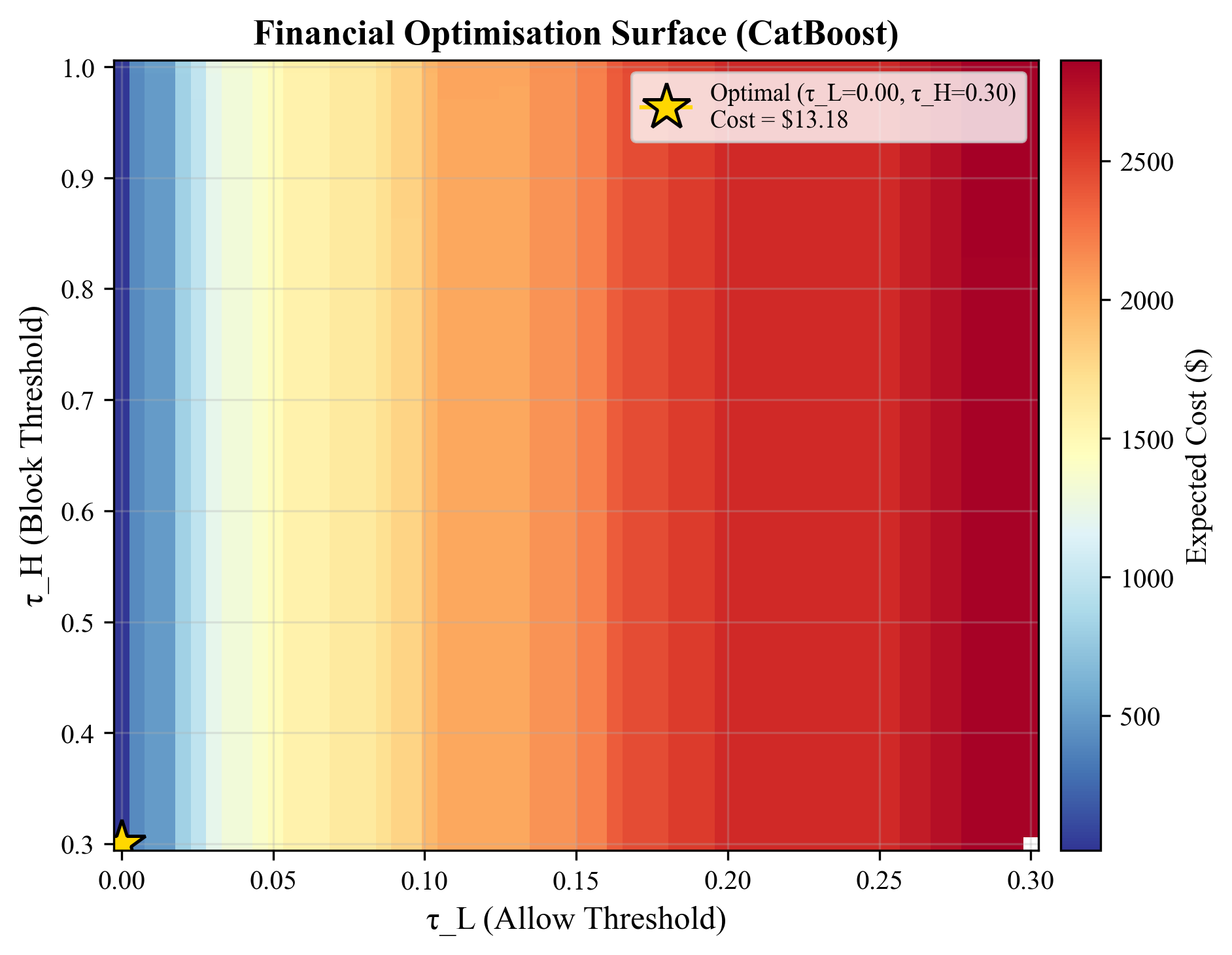}
\caption{Financial Optimisation Surface. The ``Valley of Safety'' (dark blue region) minimises total financial loss across threshold combinations.}
\label{fig:cost}
\end{figure}

Figure~\ref{fig:cost} illustrates the financial optimisation landscape. The optimal policy is $\tau_L \approx 0$, maximising threat capture while minimising false positives.

The optimal threshold ($\tau_L \approx 0$) maximises financial utility under fixed cost assumptions. However, operational deployment introduces analyst workload constraints and alert fatigue risks not captured in this model. Organisations processing 100K+ emails daily may require higher thresholds to maintain sustainable review queues, trading marginal detection gains for operational feasibility.

\subsection{GPU vs.\ CPU Infrastructure Trade-offs}
Hardware configuration specifically influences inference latency on Tesla T4 GPU infrastructure:
\begin{itemize}
    \item \textbf{DistilBERT:} 7.403~ms per email (batch size 1, GPU)
    \item \textbf{CatBoost:} 0.855~ms per email (CPU)
    \item \textbf{TF--IDF+LogReg:} 0.067~ms per email (CPU)
\end{itemize}

For organisations processing 100,000 emails per hour:
\begin{itemize}
    \item DistilBERT infrastructure: approximately 205 GPU-hours per year (annual cost: \$150--300 on cloud platforms)
    \item CatBoost infrastructure: approximately 25 CPU-hours per year (negligible cost)
\end{itemize}

\section{Conclusion}
\label{sec:conclusion}

We refine a deployment-facing comparison for BEC detection by combining cost-sensitive decision-making with adversarial robustness and latency measurement.
DistilBERT represents the high-accuracy semantic option under GPU inference (AUC 1.0000, F1 0.9981, 7.403~ms), while CatBoost provides lower-latency forensic detection under constrained compute (AUC 0.9860, F1 0.9382, 0.855~ms).
Camera-ready additions---classical baselines (TF--IDF+LogReg achieving AUC 1.0000; char $n$-gram+SVM achieving AUC 1.0000), $\psi$ ablation (marginal but theoretically motivated contribution), and homoglyph-map sensitivity (CatBoost recall drops 42\% without normalisation)---convert the study from a two-model comparison into an operationally grounded trade-off analysis.
Future work must validate performance on longitudinal operational data to address synthetic bias and temporal drift. Psycholinguistic feature constructs require empirical validation through controlled deception studies.

\section*{Acknowledgements}
We acknowledge Tim Schneller, Kofi Osei, Kwabena Kissiedu, Davis Opoku, Ephraim Abotsi, Papa Adarkwa, and Kwadwo Amanqua for feedback and support.

\textit{AI disclosure (IEEE policy):} The authors used a generative AI language model (OpenAI ChatGPT/GPT-4) to assist with language editing and clarity across the manuscript; all technical content, numerical results, and conclusions were verified and are the authors' responsibility \cite{ieee_ai_ack,openai_gpt4}.

\balance

\appendices

\section{Feature Definitions}
\label{app:features}

Table~\ref{table:features_app} lists the psycholinguistic features used in the analysis. It defines each feature and states what linguistic attribute it captures.

\begin{center}
\captionof{table}{Psycholinguistic Feature Definitions}
\vspace{0.2cm}
\label{table:features_app}
\begin{tabular}{l p{5cm}}
\toprule
\textbf{Feature} & \textbf{Description} \\
\midrule
$\psi$ Score & Interaction of politeness and urgency. \\
Authority & Count of organizational titles and imperative cues. \\
Hedges & Frequency of tentative expressions. \\
Sentiment $\Delta$ & Shift in sentiment polarity across the email. \\
Caps Ratio & Proportion of uppercase characters. \\
\bottomrule
\end{tabular}
\end{center}

\end{document}